\documentclass{sig-alternate-05-2015}
\usepackage[]{graphicx}
\usepackage{tabularx}
\usepackage{amssymb,amsmath}
\usepackage{algorithm}
\usepackage[noend]{algpseudocode}
\usepackage{booktabs}
\usepackage{multirow}
\usepackage{enumitem}
\usepackage[dvipsnames]{xcolor}
\usepackage{colortbl}
\usepackage{threeparttable}

\newcommand{\alg}[1]{Alg.~\ref{alg:#1}}
\newcommand{\fig}[1]{Fig.~\ref{fig:#1}}
\definecolor{MyPurple}{rgb}{0.7725490196078432, 0.6901960784313725, 0.8352941176470589}
\definecolor{MyPink}{rgb}{0.9686274509803922, 0.7137254901960784, 0.8235294117647058}
\definecolor{MyOrange}{rgb}{1.0, 0.7333333333333333, 0.47058823529411764}
\makeatletter
\def\@copyrightspace{\relax}
\makeatother
\begin{document}






%

\title{Incremental Factorization Machines for Persistently Cold-starting Online Item Recommendation}
%
%
%
%
%

\numberofauthors{1} 
%

\author{
%
%
\alignauthor
Takuya Kitazawa\\
       \affaddr{Graduate School of Information Science and Technology}\\
       \affaddr{The University of Tokyo, Japan}\\
       \email{k.takuti@gmail.com}
}

\maketitle
\begin{abstract}
  Real-world item recommenders commonly suffer from a persistent cold-start problem which is caused by dynamically changing users and items. In order to overcome the problem, several context-aware recommendation techniques have been recently proposed. In terms of both feasibility and performance, factorization machine (FM) is one of the most promising methods as generalization of the conventional matrix factorization techniques. However, since online algorithms are suitable for dynamic data, the static FMs are still inadequate. Thus, this paper proposes incremental FMs (iFMs), a general online factorization framework, and specially extends iFMs into an online item recommender. The proposed framework can be a promising baseline for further development of the production recommender systems. Evaluation is done empirically both on synthetic and real-world unstable datasets.
\end{abstract}

%
%


%
%

%
%



\section{Introduction}
\label{sec:introduction}

In the real-world applications such as e-commerce and online ad, a user's activity is not frequent, and item properties change dynamically over time. In a context of item recommendation, such scenario is referred to as \textit{persistent cold-start}. For instance, Booking.com \cite{Bernardi2015} shows an example of users' rare activity and mixed personas (\textit{\textbf{user-side} persistent cold-start}), and Rakuten GORA \cite{Robin2015} demonstrates price fluctuation and short life-span of packaged items (\textit{\textbf{item-side} persistent cold-start}).

Most importantly, classical recommendation techniques have some drawbacks under the persistent cold-start setting. In fact, matrix factorization (MF) \cite{Koren2009} is one of the most typical and promising techniques, but MF only holds latent vectors for every user/item IDs; there is no way to make meaningful recommendation under an unforeseen condition. By contrast, context-aware recommender systems have been recently studied in order to profile more essential users' preferences with auxiliary features. In particular, factorization machines (FMs) \cite{Rendle2012-1} are alternative effective factorization models which enable us to make context-aware recommendation with flexible feature representation. Since the persistent cold-start problem commonly occurs in real applications, high feasibility of FMs is attractive compared to more specific methods developed by industrial researchers. However, the captured context by FMs is still static, and thus FM is incomplete in terms of robustness against persistently cold-starting data.

\begin{figure}[htbp]
  \centering
  \includegraphics[width=0.8\linewidth]{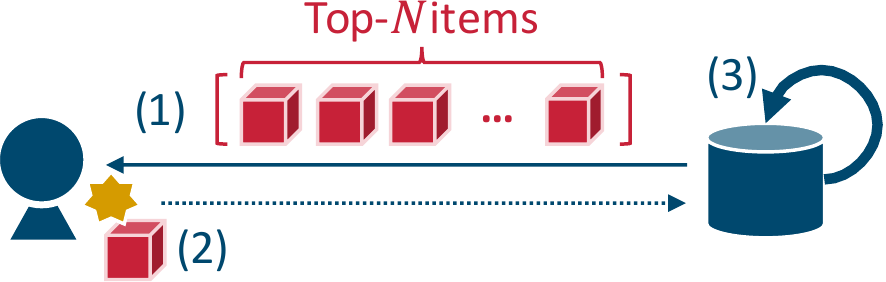}
  \vspace*{-5pt}
  \caption{\textbf{Online item recommender}: \textnormal{(1) recommend a top-$N$ list to a user, (2) interact with an item (e.g. click, buy, rate), and (3) update parameters based on the interaction.}}
  \label{fig:stream-recommender}
\end{figure}

In order to adjust the model parameters according to variation of context, this paper extends FMs into online algorithms. It should be noticed that the persistent cold-start problem is closely related to \textit{concept drift}, a phenomenon of ``the relation between the input
data and the target variable changes over time'' \cite{Gama2014} in data streams. On the user-side, since users' interests may be changed, systems must recommend different items even for the same user. Meanwhile, due to the instability of item trends and properties, one item can be preferred by totally different users at a different point in time. Past studies proved that online algorithms are effective to the unpredictable phenomena \cite{Gama2014}, so the author assumes that incremental update of FMs yields better accuracy compared to the static recommenders. In practice, online recommender systems behave as illustrated in \fig{stream-recommender}.

More concretely, the author generalizes prior work in incremental MF (iMF) \cite{Vinagre2014-1}, an online extension of a MF-based item recommender, to incremental FMs (iFMs). Here, the framework specially equips the following properties:
\begin{itemize}[noitemsep]
\item \textbf{Positive-only feedback}: model update is fast thanks to an one-pass online learning scheme.
\item \textbf{Incremental adaptive regularization}: regularization parameters are automatically adjusted on-the-fly.
\end{itemize}

\section{Components}

This section introduces the different factorization techniques which compose the proposed framework.

\subsection{Incremental Matrix Factorization}
\label{sec:iMF}

Vinagre et~al. \cite{Vinagre2014-1} proposed an efficient iMF algorithm for item recommendation, which is achieved by solving MF with a unique target value $y=1$ over the stochastic gradient descent (SGD) optimization. Moreover, they evaluated the method in a \textit{test-then-learn} scheme as outlined in \alg{prequential}. For each pair of a user $u \in U$ and an item $i \in I$, evaluation is first launched before updating the parameters.

\begin{algorithm}
\caption{Outline of the \textit{test-then-learn} procedure}
\label{alg:prequential}
\begin{algorithmic}[1]
  \Require data stream or finite set of positive events $S_+$,
  \Statex\hspace{\algorithmicindent}size of recommendation list $N$, window size $T$
  \State \underline{\textbf{Step 0:} initialize parameters}
  \State \underline{\textbf{Step 1:} batch training by using $S_0 \subset S_+$}
  \For{$(u, i) \in S_+ \setminus S_0$}
  \State \underline{\textbf{Step 2:} recommend and evaluate}
  \Statex\hspace{\algorithmicindent}$L := \{i^* \mid \textrm{$i^*$ is in top-$N$ recommended items for $u$}\}$
  \Statex\hspace{\algorithmicindent}$\mathrm{recall@}N = 1$ if $i \in L$, otherwise $0$
  \Statex\hspace{\algorithmicindent}$\mathrm{recall@}N/T := $ avg. recall@$N$ for latest $T$ samples
  \State \underline{\textbf{Step 3:} update}
  \EndFor
\end{algorithmic}
\end{algorithm}

More specifically, iMF incrementally factorizes a binary matrix $R \in \mathbb{R}^{|U| \times |I|}$ into $P \in \mathbb{R}^{|U| \times k}$ and $Q \in \mathbb{R}^{|I| \times k}$ as:
\begin{itemize}[noitemsep]
   \vspace*{-4pt}
\item \underline{\textbf{Step 1:}} Learn $P$ and $Q$ as the standard MF
\item \underline{\textbf{Step 2:}} Score items by $Q \ \mathbf{p}_u \in \mathbb{R}^{|I|}$, and \\
  \ \ \ \ \ \ \ \ \ \ \ \ recommend $N$ closest items to $1$
\item \underline{\textbf{Step 3:}} $\mathbf{p}_u \leftarrow \mathbf{p}_u + 2 \eta \ \left(\left(1 - \mathbf{p}_u^{\mathrm{T}} \mathbf{q}_i\right) \ \mathbf{q}_i - \lambda \ \mathbf{p}_u\right)$ \\
  \ \ \ \ \ \ \ \ \ \ \ \ $\mathbf{q}_i \leftarrow \mathbf{q}_i + 2 \eta \ \left(\left(1 - \mathbf{p}_u^{\mathrm{T}} \mathbf{q}_i\right) \ \mathbf{p}_u - \lambda \ \mathbf{q}_i\right)$
   \vspace*{-4pt}
\end{itemize}
where $\mathbf{p}_u, \mathbf{q}_i \in \mathbb{R}^k$ are respectively a user, item latent vector. The vectors are updated with a regularization parameter $\lambda$ and a learning rate $\eta$.

\subsection{Factorization Machines}
\label{sec:FMs}

FMs \cite{Rendle2012-1} have been recently developed as a general predictor. For an input vector $\mathbf{x} \in \mathbb{R}^d$, let us first imagine a linear model parameterized by $w_0 \in \mathbb{R}$，$\mathbf{w} \in \mathbb{R}^d$. In addition, by incorporating interactions of the $d$ input variables, the linear model is extended to FMs as:
\vspace*{-2.5pt}
\begin{equation}
\hat{y}(\mathbf{x}) := \underbrace{w_0}_{\textbf{global bias}} + \underbrace{\mathbf{w}^{\mathrm{T}} \mathbf{x}_{ }}_{\textbf{linear}} + \sum_{i=1}^d \sum_{j=i}^d \underbrace{\mathbf{v}_i^{\mathrm{T}} \mathbf{v}_j}_{\textbf{interaction}} x_i x_j, \nonumber
\label{eq:FMs}
 \vspace*{-2.5pt}
\end{equation}
where $V \in \mathbb{R}^{d \times k}$ is a rank-$k$ matrix which has $\mathbf{v}_1, \cdots, \mathbf{v}_d \in \mathbb{R}^k$. Arbitrary feature representation $\mathbf{x}$ (e.g. concatenation of one-hot vectors for several categorical variables) work well with FMs, and MF is actually a subset of the predictor.

\section{Incremental Factorization Machines}

At the beginning, the general iFM is proposed in Sec.~\ref{sec:FMs-model} and \ref{sec:FMs-reg}. Next, Sec.~\ref{sec:iFMs-pos} optimizes it for positive-only-feedback-based online item recommendation.

\subsection{General Incremental Predictor}
\label{sec:FMs-model}

This paper focuses on running FMs in an incremental fashion. Generally, learning FM requires a set of parameters $\Theta = \{w_0, \mathbf{w}, V\}$ and a loss function $\ell(\hat{y}(\mathbf{x} \mid \Theta), y)$, and the parameters can be optimized by SGD. Specifically, for a set of samples $S$, the parameters of FM are updated as \alg{FMs-model}. For simplicity, $\ell(\hat{y}(\mathbf{x} \mid \Theta), y)$ is written as $\ell$.

\newpage

\begin{algorithm}
\caption{SGD update for the model parameters of FM}
\label{alg:FMs-model}
\begin{algorithmic}[1]
  \Require $S$, learning rate $\eta$,
  \Statex\hspace{\algorithmicindent}regularization parameters $\lambda_0, \lambda_{\mathbf{w}}, \lambda_{V_1}, \dots, \lambda_{V_k}$
  \Repeat
  \For{$(\mathbf{x}, y) \in S$}
  \Statex\hspace{\algorithmicindent}\hspace{\algorithmicindent}\underline{\textbf{$\Theta$-update}}
  \State $w_0 \leftarrow w_0 - \eta \ (\frac{\partial}{\partial w_0}\ell + 2 \lambda_0 w_0)$
  \For{$i \in \{1,\dots,d\} \wedge x_i \neq 0$}
  \State $w_i \leftarrow w_i - \eta \ (\frac{\partial}{\partial w_i}\ell + 2 \lambda_{\mathbf{w}} w_i)$
  \For{$f \in \{1,\dots,k\}$}
  \State $v_{i,f} \leftarrow v_{i,f} - \eta \ (\frac{\partial}{\partial v_{i,f}}\ell + 2 \lambda_{V_f} v_{i,f})$
  \EndFor
  \EndFor
  \EndFor
  \Until{$w_0, \mathbf{w}, V$ are successfully learnt}
\end{algorithmic}
\end{algorithm}

Notice that \underline{\textbf{Step 3}} of iMF in Sec.~\ref{sec:iMF} is SGD update for single sample, so the basic idea of this paper is that we replace the step with \underline{\textbf{$\Theta$-update}} in \alg{FMs-model}. As a result, iFM which can be evaluated in the \textit{test-then-learn} scheme is derived without loss of generality.


\subsection{Incremental Adaptive Regularization}
\label{sec:FMs-reg}

Rendle \cite{Rendle2012-2} proposed an adaptive regularization scheme for FMs. As shown in \alg{FMs-reg}, the technique adjusts the regularization parameters by using a sample $(\mathbf{x}', y')$ in a validation set $S'$, an extra set of samples which is different from $S$.

\begin{algorithm}
\caption{Update $\lambda_0, \lambda_{\mathbf{w}}, \lambda_{V_1}, \dots, \lambda_{V_k}$}
\label{alg:FMs-reg}
\begin{algorithmic}[1]
  \Statex \underline{\textbf{$\lambda$-update}} using $(\mathbf{x}', y')$ sampled from $S'$
  \State $\lambda_0 \leftarrow \mathrm{max}(0, \ \lambda_0 - \eta \ \frac{\partial}{\partial \lambda_0}\ell')$
  \State $\lambda_{\mathbf{w}} \leftarrow \mathrm{max}(0, \ \lambda_{\mathbf{w}} - \eta \ \frac{\partial}{\partial \lambda_{\mathbf{w}}}\ell')$
  \For{$f \in \{1,\dots,k\}$}
    \State $\lambda_f \leftarrow \mathrm{max}(0, \ \lambda_f - \eta \ \frac{\partial}{\partial \lambda_f}\ell')$
  \EndFor
\end{algorithmic}
\end{algorithm}

\begin{figure}[htbp]
  \centering
  \includegraphics[width=0.75\linewidth]{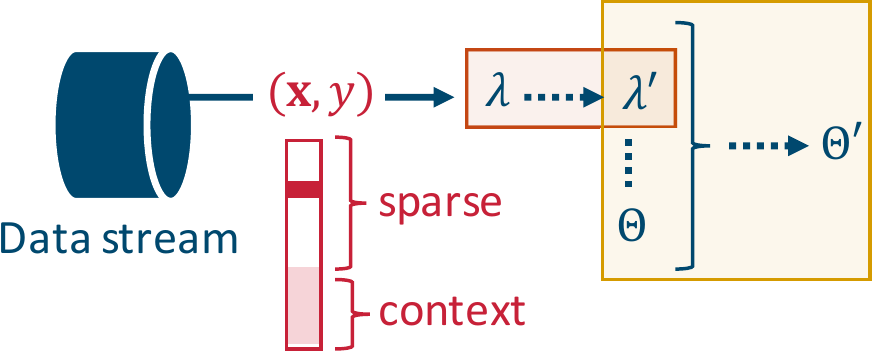}
 \vspace*{-5pt}
  \caption{\textbf{Incremental adaptive regularization.}}
  \label{fig:adapt-then-learn}
\end{figure}

Normally, \alg{FMs-reg} is launched after \underline{\textbf{$\Theta$-update}} in \alg{FMs-model}. However, when we consider incremental adaptive regularization, there is a difficulty that the validation set $S'$ will be gradually outdated in a streaming environment. A key idea to conquer the problem is that a newly observed sample $(\mathbf{x}, y)$ is handled as a \textit{pseudo} validation sample, so the regularization parameters are adjusted before updating $\Theta$ as demonstrated in \fig{adapt-then-learn}.

\vspace{-5pt}

\subsection{Context-aware Online Item Recommendation with Positive-only Feedback}
\label{sec:iFMs-pos}

This section considers a particular combination of an output $y$ and a loss function $\ell$ to utilize iFMs for online item recommendation. As the author explained in Sec.~\ref{sec:iMF}, iMF actually solves MF with $y=1$. Similarly to the approach, let us again consider the unique output for a set of positive events $S_+$. As a consequence, for a sample $(\mathbf{x}, 1) \in S_+$, our loss function is defined as: $\ell(\hat{y}(\mathbf{x} \mid \Theta), 1) = (\hat{y}(\mathbf{x} \mid \Theta) - 1)^2$.

Eventually, for arbitrary design of a feature vector $\mathbf{x}$, the proposed iFM-based item recommender which is feasible in a streaming environment can be described in the \textit{test-then-learn} framework as:
\begin{itemize}[noitemsep]
\item \underline{\textbf{Step 1:}} Learn $w_0$, $\mathbf{w}$ and $V$ as the standard FMs
\item \underline{\textbf{Step 2:}} Predict $\hat{y}(\mathbf{x} \mid \Theta)$ for every items, \\
  \ \ \ \ \ \ \ \ \ \ \ \ and recommend $N$ closest items to $1$
\item \underline{\textbf{Step 3:}} \underline{\textbf{$\lambda$-update}} $\rightarrow$ \underline{\textbf{$\Theta$-update}} with $(\mathbf{x}, 1)$
\end{itemize}

\begin{figure}[htbp]
  \centering
  \includegraphics[width=.9\linewidth]{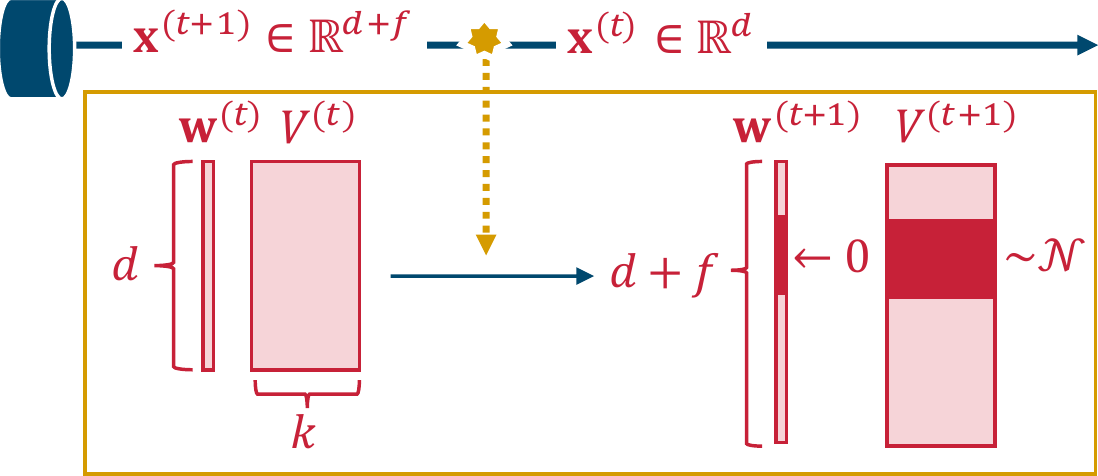}
  \vspace*{-5pt}
  \caption{\textbf{Incorporating new features into a model.} \textnormal{When dimension of $\mathbf{x}$ is increased by new users, items and/or contexts, initial values fill the corresponding parameters.}}
  \label{fig:new-feature}
\end{figure}

Importantly, since the number of users and items on real-world online applications is not constant, our systems must incorporate new users and items into a current model somehow. For example, iMF handles a new user (item) as an additional row of $P$ ($Q$) (i.e. $\mathbf{p}_{|U|+1}$ for a new user, $\mathbf{q}_{|I|+1}$ for a new item obtained from Gaussian). Hence, iFMs also take the simple approach that a zero and random vector are respectively inserted into $\mathbf{w}$ and $V$ as the initial parameters of new features. \fig{new-feature} depicts detection and insertion of new features in a stream of input vectors. It is notable that, beyond new users and items, adding new contextual variables is also possible in the middle of data streams.

\vspace*{-3pt}

In terms of computational complexity, iFMs compute the interaction term $\sum_{i=1}^d \sum_{j=i}^d \mathbf{v}_i^{\mathrm{T}} \mathbf{v}_j x_i x_j$ in $\mathcal{O}\left(kN_z(\mathbf{x})\right)$ by letting the number of nonzero elements in $\mathbf{x}$ be $N_z(\mathbf{x})$. In fact, this complexity is efficient enough due to sparsity of $\mathbf{x}$, but running time will be relatively long compared to the single vector updating of iMF. Therefore, there is trade-off between running time and context-awareness in practice.



\section{Experiments}

\subsection{Evaluation Method}

\begin{figure}[htbp]
  \centering
  \includegraphics[clip,width=0.9\linewidth]{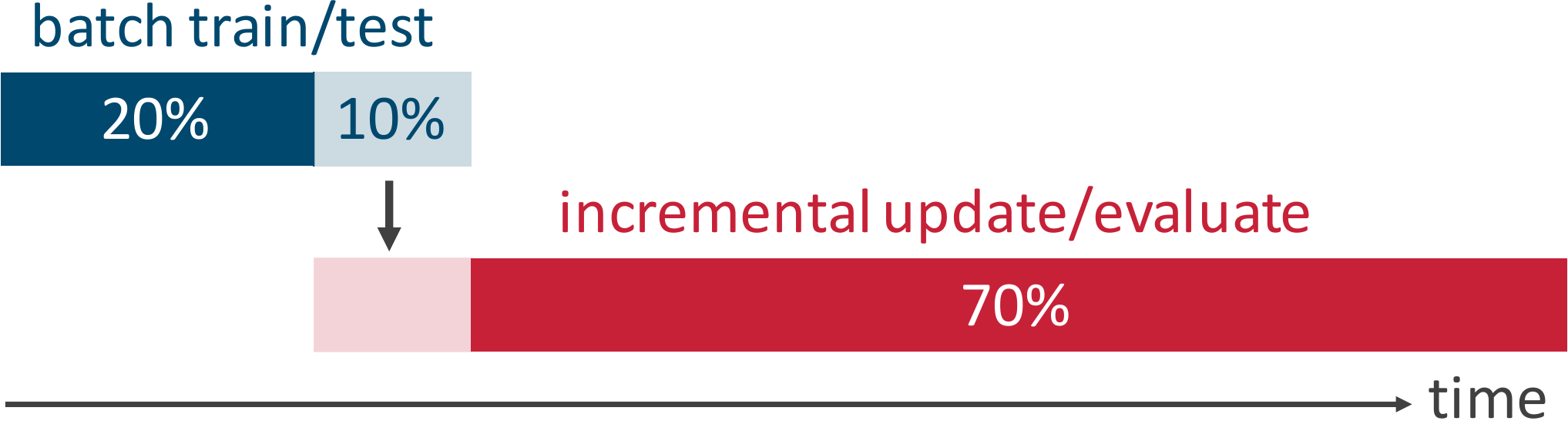}
  \vspace*{-10pt}
  \caption{\textbf{Splitting time-stamped samples.}}
  \vspace*{-10pt}
  \label{fig:sample-split}
\end{figure}

In the experiments, the \textit{test-then-learn} procedure described in Sec.~\ref{sec:iMF} is launched for time-stamped unstable datasets. As shown in \fig{sample-split}, the samples are separated similarly to what Matuszyk et~al. \cite{Matuszyk2015} did. At the beginning, a batch train/test step is executed for the first 20\% training and following 10\% validation samples. Next, the 10\% samples are just used for one-pass model updating, and the incremental evaluation step is finally performed for the remaining 70\%.

\newpage

Additionally, the 70\% samples are also evaluated by mean percentile rank (MPR) in order to assess an ordered list of items obtained from \underline{\textbf{Step 2}}. The metric is calculated based on percentile rank of the correct items in the ordered lists; that is, $\mathrm{MPR}=0\%$ indicates the best result that a recommender always gives the highest rank to a correct item, and $\mathrm{MPR}=100\%$ is opposite. In contrast to recall@$N$/$T$ which evaluates top-$N$ items, MPR can measure users' overall satisfaction from all items.

Following methods were employed as the competitors:
\begin{itemize}[noitemsep]
  \vspace*{-4pt}
  \item \textbf{static MF}: the traditional MF,
  \item \textbf{iMF}: a fast extension of MF introduced in Sec.~\ref{sec:iMF},
  \item \textbf{static FMs}: the parameters are not updated in the incremental stage similarly to \textbf{static MF}.
  \vspace*{-4pt}
\end{itemize}
The author implemented all of the methods in Python 3.5.1, and the code was run in a typical personal computer with the 2.7 GHz Intel\textregistered \ Core\texttrademark \ i7 CPU and 4GB RAM.

Datasets used in the experiments were binarized version of MovieLens 100k (ML100k)\footnote{\url{http://grouplens.org/datasets/movielens/}} and synthetic click data; the former is a real-world example of user-side volatility, and the latter shows item-side instability. Table~\ref{tab:data} summarizes statistics of the data. Each pair of a method and a dataset was tested five times with different initial parameters.

\vspace*{-15.pt}
\begin{table}[htb]
  \centering
  \caption{Statistics of the datasets.}
  \begin{threeparttable}
  \begin{tabular}{|l|r|r|r|r|r|} \hline
    Dataset & Users & Items & \multicolumn{3}{c|}{Positive events} \\ \cline{4-6} & & & 20\% & 10\% & 70\% \\ \hline \hline
    ML100k & 928 & 1172 & 4240 & 2120 & 14841 \\ 
    Synthetic & 3570\tnote{$\dagger$} & 5 & 714 & 357 & 2499 \\ \hline
  \end{tabular}
  \begin{tablenotes}\footnotesize
    \item[$\dagger$] Random demographics were generated instead of user ID itself, so \# of users and clicks are same.
  \end{tablenotes}
  \end{threeparttable}
  \label{tab:data}
\end{table} 
\vspace*{-10.pt}

\subsection{Results and Discussions}

\begin{table*}[htb]
  \centering
  \caption{Hyperparameters and average experimental results of 5 trials. \textnormal{The best accuracy is written in \textbf{bold}.}}
  \begin{tabular}{|c|c|c|c|c|c|c|} \hline
    Dataset & Method & Hyperparameters & \multicolumn{2}{c|}{Running time [sec.]} & recall@$N$/$T$ & MPR [\%] \\ \cline{4-5}
    & & & recommend & update & \textit{mean} ($\pm$ \textit{std}) & \textit{mean} ($\pm$ \textit{std}) \\ \hline \hline
    
    & static MF & $k=40$, $\eta=0.002$ & 0.00016 & --- & 0.021 ($\pm$ 0.004) & 49.68 ($\pm$ 0.320)  \\
    \textbf{ML100k} & iMF & $\lambda=0.01$ & 0.00015 & 0.00003 & 0.026 ($\pm$ 0.006) & 47.32 ($\pm$ 0.359) \\ \cline{2-3}
    (@10/3000) & static FMs & $k=40$, $\eta=0.004$ & 0.02427 & --- & 0.023 ($\pm$ 0.008) & 36.07 ($\pm$ 0.127) \\
    & iFMs & $\lambda_0=2.0$,  $\lambda_{\mathbf{w}}=8.0$, $\lambda_{V_k} = 16.0$ & 0.02449 & 0.00150 & \textbf{0.035} ($\pm$ 0.008) & \textbf{32.55} ($\pm$ 0.023) \\ \hline
    
    & static MF & $k=2$, $\eta=0.0003$ & 0.00002 & --- & 0.271 ($\pm$ 0.248) & 54.59 ($\pm$ 4.386) \\
    \textbf{Synthetic} & iMF & $\lambda=0.01$ & 0.00002 & 0.00003 & \textbf{0.316} ($\pm$ 0.213) & 49.24 ($\pm$ 2.254) \\ \cline{2-3}
    (@1/500) & static FMs & $k=2$, $\eta=0.00006$ & 0.00325 & --- & 0.271 ($\pm$ 0.248) & 37.83 ($\pm$ 3.385) \\
    & iFMs & $\lambda_0=\lambda_{\mathbf{w}}=\lambda_{V_k} = 0.01$ & 0.00315 & 0.00026 & \textbf{0.316} ($\pm$ 0.208) & \textbf{34.26} ($\pm$ 1.429) \\ \hline
  \end{tabular}
  \label{tab:result}
\end{table*}

\noindent \textbf{ML100k.} Since we focus on item recommendation, ML100k was binarized by extracting 5-starred rating events. An input vector of iFMs was designed as:
\begin{eqnarray*}
&& \mathbf{x} = ( \
  \underbrace{\textrm{\colorbox{MyOrange!50}{\makebox[4em]{\strut user ID}}},}_{1/|U|}
  \underbrace{\textrm{\colorbox{MyOrange!50}{\makebox[6em]{\strut demographics}}},}_{3/23} 
  \underbrace{\textrm{\colorbox{MyPink!50}{\makebox[4em]{\strut movie ID}}},}_{1/|I|} \\
  &&
  \underbrace{\textrm{\colorbox{MyPink!50}{\makebox[3em]{\strut genre}}},}_{18/18} 
  \underbrace{\textrm{\colorbox{MyPink!50}{\makebox[7em]{\strut last rated genre}}},}_{18/18}
  \underbrace{\textrm{\colorbox{MyPurple!50}{\makebox[2em]{\strut day}}},}_{1/7}
  \underbrace{\textrm{\colorbox{MyPurple!50}{\makebox[6em]{\strut last rated day}}}}_{1/7} 
\ ).
\label{eq:vector_ml}
\end{eqnarray*}
``demographics'' includes user's occupation (1/21), sex (1/1) and age (1/1), and ``day'' is day of week in the timestamps. Note that the numbers below the underbraces indicate \{\textit{max. \# of nonzero dimensions}\}/\{\textit{\# of total dimensions}\}.

Most users on ML100k report positive events just for the initial rating activity, and they do not rate any more. Even if some users continuously rate movies, intuition tells us that their interests change over time. Thus, ML100k is a real-world example of user-side persistent cold-start. 

\begin{figure}[htbp]
  \centering
  \includegraphics[clip,width=\linewidth]{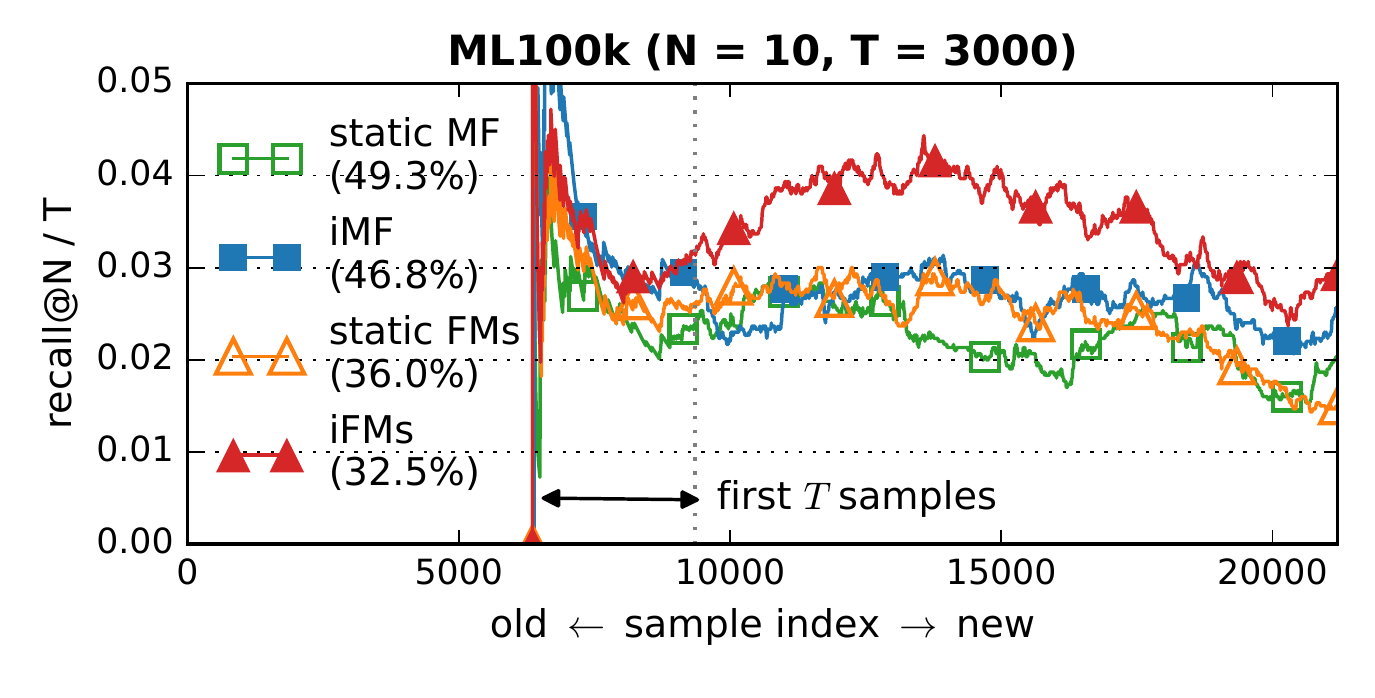}
 \vspace*{-25pt}
  \caption{\textbf{Recall behavior on ML100k.} \textnormal{Higher recall is better on the y-axis. MPR is written in the legend area.}}
  \vspace*{-15pt}
  \label{fig:result_ml100k}
\end{figure}

\fig{result_ml100k} shows the best recall behaviors on ML100k obtained from the five \textit{test-then-learn} trials. ML100k is rich in both user and item features, and we also considered time-related contexts. Consequently, static FMs and iFMs respectively outperformed static MF and iMF, especially in terms of MPR. In addition, it is clear that online recommenders (iMF and iFMs) worked effectively compared to the static counterparts.

\noindent \textbf{Synthetic.} Synthetic click data was generated as an example of item-side persistent cold-start, based on a rule-based procedure demonstrated in \cite{Aharon2013}. In particular, our generator first produced 0.5 million impressions of five ad variants, and additional half million impressions were also generated after updating a rule for the most popular ad. As a result, from the one million impressions, 3,570 synthetic clicks were observed as positive events with erratic trend. Here, $\mathbf{x}$ for the synthetic data was:
\begin{eqnarray*}
& \mathbf{x} = ( \
  \underbrace{\textrm{\colorbox{MyOrange!50}{\makebox[2em]{\strut age}}},}_{1/1} 
  \underbrace{\textrm{\colorbox{MyOrange!50}{\makebox[2em]{\strut sex}}},}_{1/1}
  \underbrace{\textrm{\colorbox{MyOrange!50}{\makebox[5em]{\strut geo (state)}}},}_{1/50}
  \underbrace{\textrm{\colorbox{MyPink!50}{\makebox[3em]{\strut ad ID}}},}_{1/|I|}
  \underbrace{\textrm{\colorbox{MyPink!50}{\makebox[4em]{\strut category}}}}_{1/3}
\ ).
\label{eq:vector_ad}
\end{eqnarray*}

\begin{figure}[htbp]
  \centering
  \includegraphics[clip,width=\linewidth]{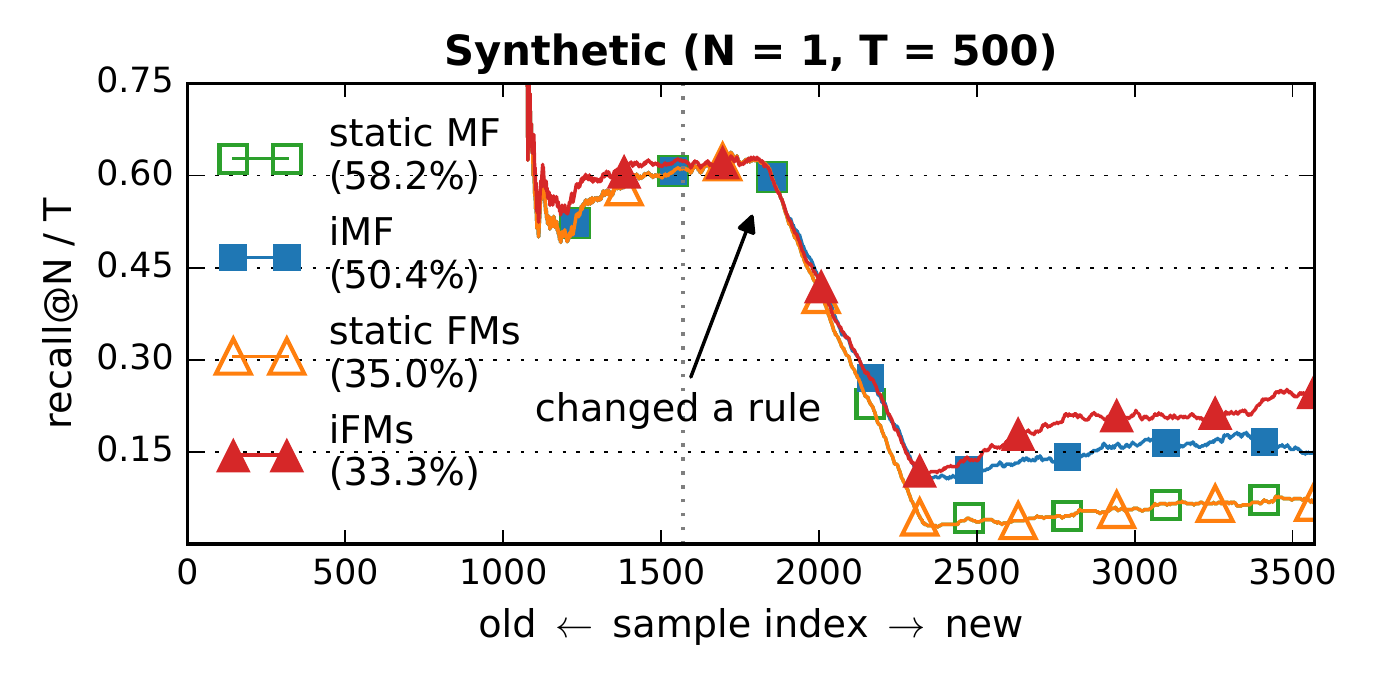}
 \vspace*{-25pt}
  \caption{\textbf{Recall behavior on synthetic click data.}}
  \label{fig:result_synthetic}
\end{figure}

The best result on the synthetic data is illustrated in \fig{result_synthetic}. All methods were easily fit to the batch training samples due to the simplicity of data, but the difference can be observed after the most popular ad was changed. While recall of the static methods declined significantly, iMF and iFMs evidently adapted to the variation as expected. Moreover, MPR and recall of iFMs were even better than iMF. It should be noted that, even though static MF and FMs demonstrated the similar recall behavior, FMs showed reasonably lower MPR compared to MF.

\vspace*{-10pt}

Finally, Table~\ref{tab:result} summarizes the results of 5 trials. The positive effect of context-awareness and online model updating was validated in terms of accuracy. On the other hand, running time of FMs was more than 100 times slower than MF. iFMs updated the parameters in a millisecond range, and recommendation for a user was done at least 30 milliseconds. iFM thus seems to satisfy practical time requirements for now, but higher-dimensional and denser input vectors may lead worse results in the future. As the author mentioned in Sec.~\ref{sec:iFMs-pos}, the fact proved trade-off between context-awareness and efficiency.

Overall, the proposed online item recommender based on iFMs worked well as generalization of iMF. In case that an organization develops production recommender systems, our highly feasible framework can be an easy-to-implement baseline. Furthermore, availability of many third-party libraries (e.g. \cite{Bayer2015,Rendle2012-1}) is an important advantage of FMs.

\section{Conclusion}
This paper has proposed an iFM-based context-aware online item recommender. Experimental results have demonstrated not only feasibility and effectiveness of the technique but also a new challenge in computational efficiency.

%

\end{document}